# A Fuzzy Expert System for Earthquake Prediction, Case Study: The Zagros Range

Arash Andalib, Mehdi Zare, Farid Atry

National Center for Earthquake Prediction
International Institute of Earthquake Engineering and Seismology

**ABSTRACT**

A methodology for the development of a fuzzy expert system (FES) with application to earthquake prediction is presented. The idea is to reproduce the performance of a human expert in earthquake prediction. To do this, at the first step, rules provided by the human expert are used to generate a fuzzy rule base. These rules are then fed into an inference engine to produce a fuzzy inference system (FIS) and to infer the results. In this paper, we have used a Sugeno type fuzzy inference system to build the FES. At the next step, the adaptive network-based fuzzy inference system (ANFIS) is used to refine the FES parameters and improve its performance. The proposed framework is then employed to attain the performance of a human expert used to predict earthquakes in the Zagros area based on the idea of coupled earthquakes. While the prediction results are promising in parts of the testing set, the general performance indicates that prediction methodology based on coupled earthquakes needs more investigation and more complicated reasoning procedure to yield satisfactory predictions.

## 1. INTRODUCTION

Iran is vulnerable to a catastrophic earthquake and the occurrence of such an earthquake with a magnitude of 7 on the Richter scale is mostly probable. Among many regions in Iran with faults that threat the country, the Zagros belt in south-southwest is the most seismically active region in the Iranian plateau, and unfortunately, in case of a major earthquake, we will witness tragic losses. Therefore, a reliable earthquake alarm system could be of extremely great importance.

The current improvements in seismology have brought us invaluable knowledge about earthquakes. Owing to both earthquake engineering methods and the modern equipment, now we have achieved precise data of this phenomenon. This is very useful for the artificial intelligence-based alarm systems that need a huge amount of data to be designed. Among AI-based learning machines, expert systems are a good candidate to be applied to the problem of earthquake prediction.

Expert systems are designed and created to facilitate tasks in different fields. These systems provide consistent answers for repetitive decisions, processes and tasks. They are able to hold and maintain significant levels of information. Furthermore, they never forget to ask a question, as a human might. The foundation of a successful expert system depends on a series of technical procedures and development that may be designed by certain technicians and related experts.

This paper is aimed to design a framework for creation a fuzzy expert system to predict major earthquakes, based on the idea of coupled earthquakes. By couple earthquakes we mean two almost contemporary major earthquakes occurred in a nearby area. Some earthquake experts believe that a couple earthquakes may be followed by another moderate or bigger earthquake. Here we are not to validate or reject this idea, but we would like to use it as the foundation of the rule base of our expert system. According to this idea, a human expert inherently follows the next three steps to raise an earthquake alarm

(1) consider the current coupled earthquakes in the area
(2) find the most powerful ones
(3) if the magnitude of the couple is greater than M and the distance between the earthquakes of the couple was less than N miles

then the occurrence of a major earthquake in the following is probable.

Obviously, there is some darkness in the above statements, including the value of M, N and the time between the two mates in the couple. A human mind inherently could surpass this vagueness, while a learning machine cannot. Fuzzy logic proposed by zadeh [1] is aimed to solve such problems. To employ the fuzzy logic, we first need to translate our crisp variables into fuzzy variables, a process called as fuzzification. At the next step, we employ one of the designed inference engines for fuzzy reasoning [2] to analyze fuzzified inputs and infer the output. Parameters used in fuzzification and fuzzy inferring are set by the human expert and a slip in these parameters cause a reduction in the performance of the whole system. An adaptive network-based fuzzy inference system (ANFIS) can solve this problem. ANFIS gets the fuzzy inference system, with values initially proposed by a human expert and then refines the membership and linear functions. For example, a human expert may believe that a couple is consisted of two earthquakes, both of them $M \geq 5$, where their location is closer than 90 miles and are not separated more than 2 months in time. In the case of these conditions are fulfilled, the human expert may conclude that an earthquake is probable in the next 6 months in the area. However, parameters used by human experts are varied from expert to expert. To design an expert system, we initially let the parameters to be taken from a wide range to covers all experts' definitions of coupling. At the second step, it is leaved to ANFIS to refine these parameters.

This paper is organized in the following manner. In Section 2, we explain about fuzzy inferring and the adaptive network-based fuzzy inference system. In Section 3, we present our results from applying our expert system to the earthquake catalogue of Zagros area to provide alarms for probable earthquakes. Section 4, concludes the work.

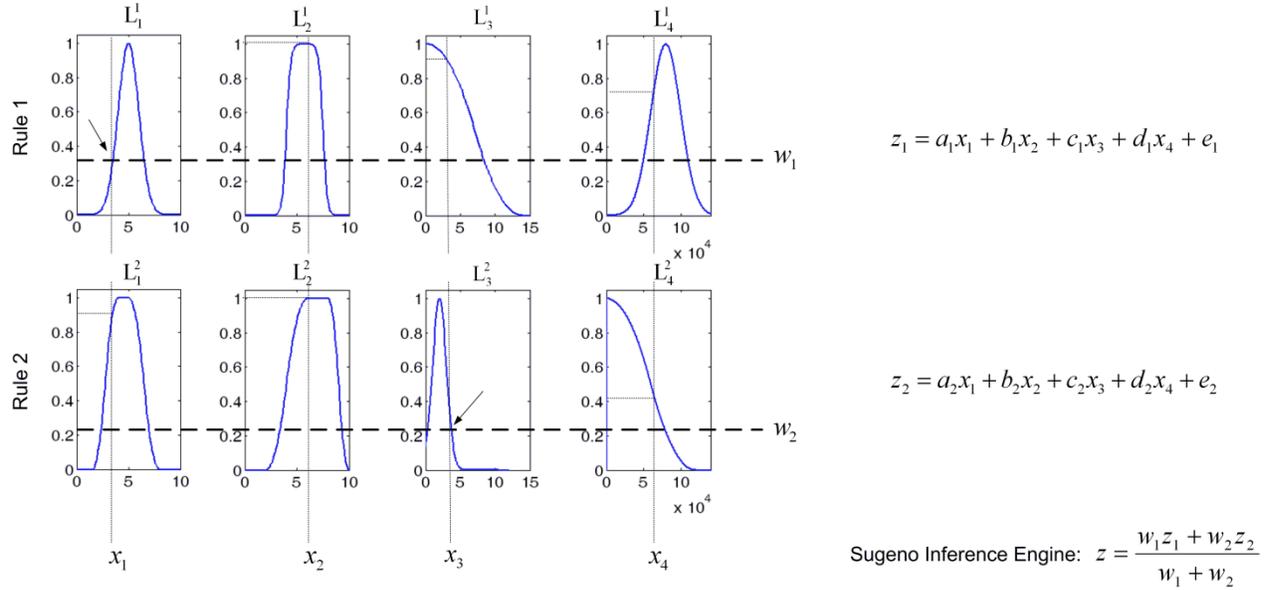

Figure 1. *Sugeno type fuzzy inference system with two rules and four membership functions.*

## 2. METHODOLOGY

While variables in mathematics usually take numerical values that are constrained to the two truth values, TRUE and FALSE, in fuzzy logic we often use linguistic variables that can take values between 0 and 1 as the degrees of truth of a statement represented by the output of a function known as membership function [1].

Membership functions take crisp values and return their degree of truth, a process known as fuzzification. In the next step, the output values are processed by an engine, called fuzzy inference, to infer the results. There are several fuzzy inference engines, such as Mamdani [3] and Sugeno [4]. Sugeno engine builds a set of rules of type "if $X$ is linguistic variable then $Y=f(X)$", where $X=[x_1, x_2, …, x_N]$, $N$ is the number of inputs and $f(X)$ is a linear function of $X$. Here, the degree of truth $w_i$ of the antecedent for rule $i$ is obtained by a combination of all its linguistic variable $L^i_j$ with AND or OR operators, where $L^i_j$ is the $j$th linguistic variable in the $i$th rule. The Sugeno type engine uses minimum or product functions as AND, and maximum or probabilistic product as OR. The antecedent's degree of truth for rule $i$ is considered as the weight of the crisp consequent of that rule $z_i$. The weighted values of $z_i$ are then averaged to produce the final output. Figure 1 shows such a system with two rules, four variables with AND operator as the connective.

Appropriate membership functions in antecedents and accurate linear functions in consequents are needed to be defined precisely. Sometimes, there is not either a human expert or membership and linear functions defined by him are not precise enough to be used in the fuzzy inference system. In 1993 Jang proposed adaptive network-based fuzzy inference system (ANFIS) [5] as a tool for refining and learning membership functions and parameters of linear functions. To do this, he estimates the parameters of linear function by least mean square (LMS) and then back propagation algorithm updates the so called membership functions.

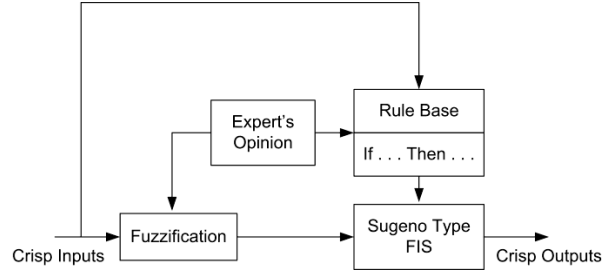

Figure 2. *The architecture of an expert based on the Sugeno type fuzzy inference system.*

The ability of learning membership functions empowers ANFIS to extract knowledge from numerical dataset and gives it more flexibility than fuzzy inference systems with membership functions preset by a human expert. In addition, this feed forward network needs lower amount of data to be trained and has better generalization performance in compare with the neural networks [5]. The architecture of the Sugeno type expert system that is used in this paper is depicted in Figure 2.

## 3. EXPERIMENTAL RESULTS

We employed our expert system to predict the major earthquakes in the Zagros range. The seismicity map of the area and the earthquakes $M \geq 5$ is illustrated in Figure 3. To set an initial fuzzy expert system, we asked a human expert used to predict the future earthquakes in Zagros area with the strategy of coupled earthquakes to give us the parameters, which he uses in his strategy. The proposed values were $M \geq 5$, three months, and 190 miles for minimum magnitude of earthquakes, maximum time distance and maximum geographical distance, respectively.

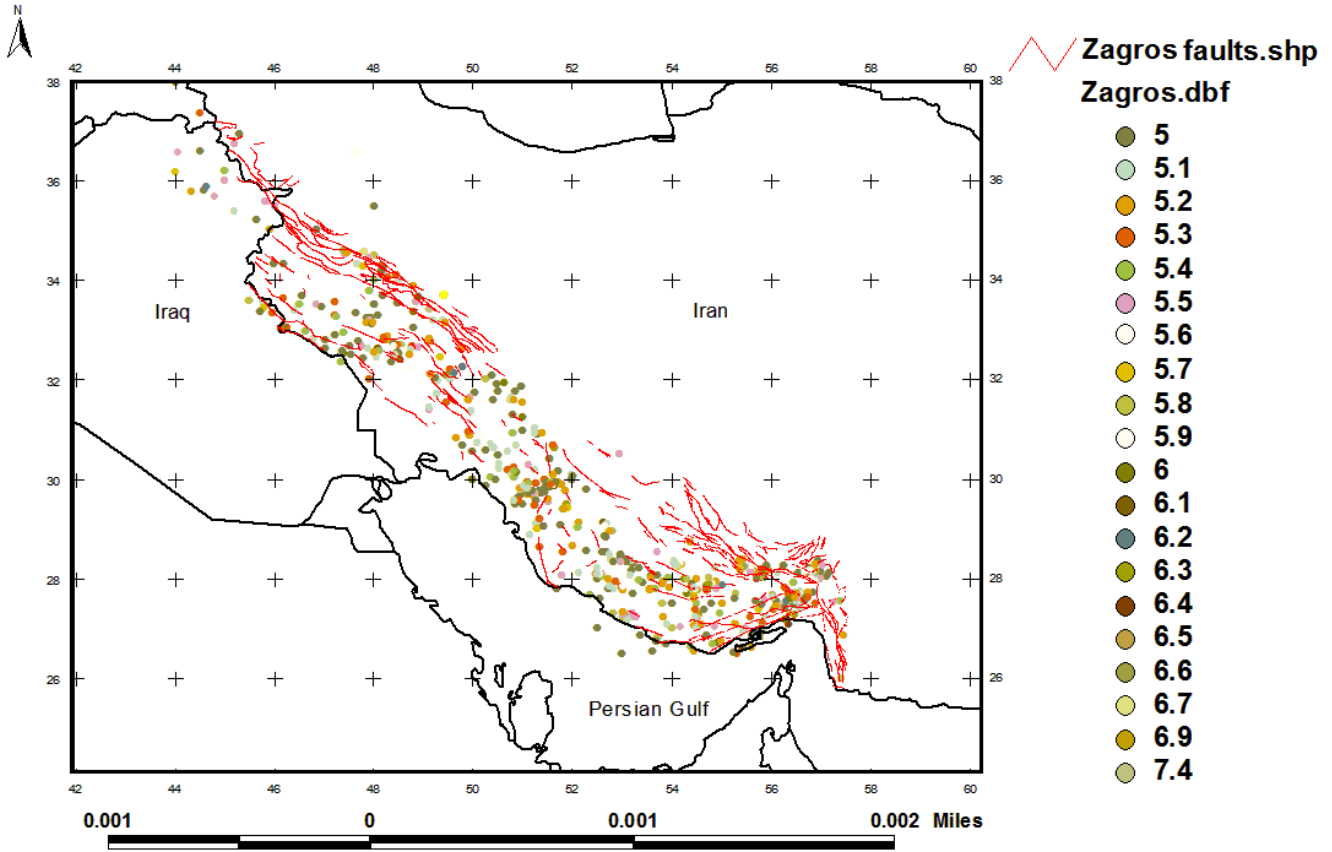

Figure 3. *Seismicity map of Zagros*

For each earthquake $x_1$, we found the biggest earthquake $x_2$ that may serve as its mate. Magnitude of these two earthquakes $x_1$ and $x_2$, their time distance $x_3$ and their geographical distance $x_4$ were the parameters stored for each couple. These parameters contain information about the next probable earthquake. For each couple the magnitude of the biggest earthquake in the next 6 months was assigned as the target during the training set, so each datum has 5 elements, 4 inputs and one output. These input variables are fed to the ANFIS. The network applies the membership functions to inputs. In each rule the two first memberships, belongs to the magnitudes of earthquakes with linguistic variables $L_1^1$ and $L_1^2$ and its mate $L_2^1$ and $L_2^2$. The next two membership functions belong to the time between two quakes in a couple, with linguistic variables $L_3^1$ and $L_3^2$ and the last memberships belong to the geographical straight distance in meter, with $L_4^1$ and $L_4^2$ as the corresponding linguistic variable.

Thereafter, ANFIS obtains the degrees of truth values $w_1$ and $w_2$ of its two rules. These degrees explain how much the rules are true and how much they should affect the final output in compare with the other rules. In the final step, the consequents $z_1$ and $z_2$ are weighted and averaged to build the final output $z$. $z$ is the magnitude of a probable earthquake in the following 6 months.

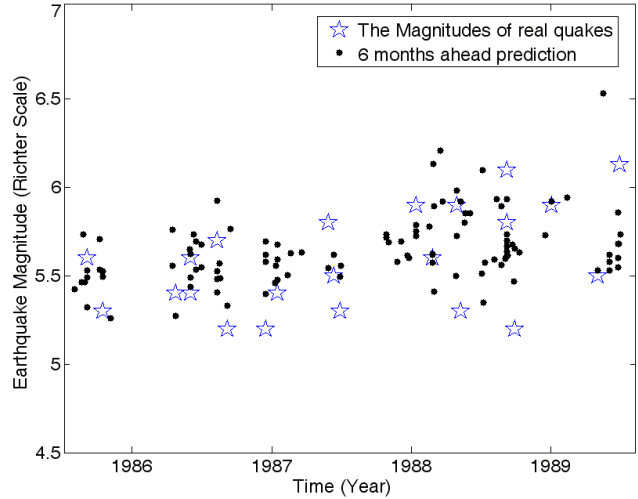

Figure 4. *Prediction results over a part of test set*

To train the ANFIS, we used the Zagros Earthquake catalogue's data before year 1985 as the training and validation sets. The time period between 1985 through 2000 is considered as the test period.

The results of applying our expert system to the prediction task in Zagros area in the period of 1985 through 1990 are presented in Figure 4. The blue stars in the figure are the actual earthquakes occurred in this period. The black dots are the 6 months ahead predictions, which are provided by the output of our expert system. The earthquake alarms just before 1986, which predict an earthquake of 5.2≤ $M$ ≤5.7 are confirmed by the occurrence of two earthquakes with the predicted time and magnitude. Obviously, there are some other cases of acceptable predictions from 1986 through the middle of 1989. The last prediction of the system in this period is very interesting, as a strong earthquake is predicted successfully about 1 month before occurrence of the quake.

Although the results of our algorithm in the above part of the test set is promising; unfortunately the predictions after 1990 are not as good as Figure 3. In the whole of the test period, there are 36 false alarms out of 147 for $M$ ≥5.5. For $M$ ≥6 the rate of false alarms approaches to %40 which is almost unacceptable. It should be mentioned that the performance of the prediction drops for strong and major earthquakes, partly because of lack of data for these range of earthquakes

## 4. CONCLUSIONS

In this paper we built a fuzzy expert system reproducing the human expert reasoning for 6 months ahead predictions. The strategy of the human expert was based on coupled earthquakes. The designed expert was then tuned by the adaptive network-based fuzzy inference system to improve its prediction accuracy. The trained expert system applied to raise earthquake alarms for the Zagros area. Although, the results are almost acceptable for earthquakes $M$ ≥5.5, however, for the earthquakes $M$ ≥6 false alarms severely increase. The general performance of our expert system indicates that the creation of such a system based on coupled earthquakes needs more investigation and more complicated reasoning procedure to yield satisfactory predictions. Furthermore, to design a reliable expert system, we need to simulate the performance of a human expert with more reliable prediction strategies.

## AKNOWLEDGEMENT

The authors would like to thank L. Mahshadnia and S. A. Hashemi for their fruitful assistance.

## 5. REFERENCES

[1] T. Takagi, M. Sugeno, "Fuzzy identification of systems and its applications to modeling and control," *IEEE Trans. Syst., Man Cyber,* vol. 15, pp 116-132, 1985.

[2] L. X. Wang, *A course in fuzzy systems and control*, 1st edition, Prentice-Hall International Inc., Upper Saddle River, 1997.

[3] E.H. Mamdani, "Applications of fuzzy logic to approximate reasoning using linguistic synthesis," *IEEE Trans on Computers*, vol. 26, no. 12, pp. 1182-1191, 1977.

[4] M. Sugeno, *Industrial applications of fuzzy control*, Elsevier Science Inc., New York, 1985.

[5] R. Jang, "ANFIS: Adaptive-network-based fuzzy inference system", *IEEE Trans. Syst. Man Cyber.,* vol. 23, no. 3, pp 665-685, May/June, 1993.

[6] L. A. Zadeh, "Outline of a new approach to the analysis of complex systems and decision processes," *IEEE Trans. Syst. Man Cyber.,* vol. 3, pp 24-44, Jan. 1973.